\documentclass{article}

\PassOptionsToPackage{numbers}{natbib}
\usepackage[final]{nips_2017}

\usepackage[utf8]{inputenc} 
\usepackage[T1]{fontenc}    
\usepackage{hyperref}       
\usepackage{url}            
\usepackage{booktabs}       
\usepackage{amsfonts}       
\usepackage{nicefrac}       
\usepackage{microtype}      
\usepackage{amsmath}
\usepackage{enumitem}

\usepackage{graphicx}
\usepackage{wrapfig}
  
\usepackage{color}

\newcommand{\justify}[1]{\mbox{\ \ #1}}

\title{Training Ensembles to Detect Adversarial Examples}

\author{
  Alexander Bagnall \\
  Ohio University \\
  Athens, OH 45701 \\
  \texttt{ab667712@ohio.edu} \\
  \And
  Razvan Bunescu \\
  Ohio University \\
  Athens, OH 45701 \\
  \texttt{bunescu@ohio.edu} \\
  \And
  Gordon Stewart \\
  Ohio University \\
  Athens, OH 45701 \\
  \texttt{gstewart@ohio.edu} \\
}

\begin{document}

\maketitle

\begin{abstract}
  We propose a new ensemble method for detecting and
  classifying adversarial examples generated by state-of-the-art
  attacks, including DeepFool and C\&W. Our method works by training
  the members of an ensemble to have low classification error on random
  benign examples while simultaneously minimizing agreement on
  examples outside the training distribution. We
  evaluate on both MNIST and CIFAR-10, against oblivious and
  both white- and black-box adversaries.
\end{abstract}

\section{Introduction}
\label{introduction}
As is now widely understood, neural networks can easily be fooled
by small but cleverly chosen input perturbations, so-called
adversarial examples~\cite{szegedy2013intriguing}.  We propose a new ensemble
learning method that enables detection -- in oblivious, black-, and
white-box settings -- of adversarial examples generated by
state-of-the-art attacks including DeepFool~\cite{moosavi2016deepfool}, the basic
iterative method~\cite{kurakin2016adversarial}, and C\&W~\cite{carlini2017towards}.

In our method, multiple instances of a base model are trained jointly
to minimize a cross-entropy loss. An additional loss term encourages
the models in the ensemble to have highly varying predictions on
examples not from the source data distribution (contrast
with~\cite{he2017adversarial}, which studies ensembles of distinct
defenses).  The ensemble labels an input as adversarial when there is
little consensus on that input. Our method is more sensitive than an
undefended model to non-adversarial (random) input perturbations but
can detect 68.1\% of the adversarial examples generated on CIFAR-10 by
C\&W~\cite{carlini2017adversarial}, an attack method that was recently
shown to be quite effective against existing defenses.

The rest of the paper is structured as follows: in
Section~\ref{sec:method} we describe a simple, computationally
inexpensive, and attack-agnostic ensemble learning method for
detecting and classifying adversarial examples; in
Section~\ref{sec:results} we experimentally evaluate ensemble
classifiers trained using our method against the adversarial examples
generated by the Fast Gradient Sign (FGS)
method~\cite{goodfellow2014explaining}, the Basic Iterative
method~\cite{kurakin2016adversarial},
DeepFool~\cite{moosavi2016deepfool}, and
C\&W~\cite{carlini2017towards}, on both MNIST and CIFAR-10. The paper
concludes with a summary of the approach and ideas for future work.

\section{Ensemble Method for Classification and Adversarial Detection}
\label{sec:method}

We propose to train an ensemble of $N$ models that label clean
examples accurately while also disagreeing on randomly perturbed
examples. At test time, the ensemble will be used for both adversarial
detection and classification: the label achieving the maximum
agreement on a test example is output, unless the agreement is too
low, in which case the example is labeled as adversarial.

Let $W_n = [{\mathbf w}_n^k]_{1 \leq k \leq K}$ be the matrix of
parameters used by the softmax layer to compute the posterior
probabilities corresponding to $K$ classes, and let $W = [W_n]_{1 \leq
  n \leq N}$ be the 3-dimensional tensor of the softmax parameters for
the entire ensemble. Let ${\mathbf \sigma}_n({\mathbf x}) =
\text{softmax}(W_n, {\mathbf x}) = [\sigma_n^k({\mathbf x})]_{1 \leq k
  \leq K}$ be the vector of softmax outputs computed by the ensemble
member $n$ on an input example ${\mathbf x}$. If ${\mathbf r}({\mathbf
  x})$ is the representation computed by the layer preceding the
softmax, then $\sigma_n^k({\mathbf x}) \propto exp({{\mathbf
    w}_n^k}^T{\mathbf r}({\mathbf x}))$. Furthermore, let
$\tilde{{\mathbf x}} = {\mathbf x} + {\mathbf{\epsilon}}$ be the
randomly perturbed version of training example ${\mathbf x}$, obtained
by adding perturbation values $\epsilon$ that are sampled uniformly at
random from $[-\eta, \eta)$, where $\eta$ is a hyperparameter that
  controls the $\mathrm{L}_{\infty}$-norm of the perturbation
  vector. To achieve the dual objective of high accuracy on clean
  examples and disagreement on randomly perturbed examples, we define
  the cost function $J = J_e + \lambda J_a$ shown in
  Equation~\ref{eq:cost} below, with two components: $J_e$ is the
  standard cross-entropy error for clean example ${\mathbf x}$ and its
  true label $y$, averaged over all ensemble members; $J_a$ is the
  mean agreement among ensemble members, where agreement between two
  members is captured through the dot product of their softmax output
  vectors.
\begin{eqnarray}
  J({\mathbf x}, y, W) & = & \;\;\quad J_e({\mathbf x}, y, W) \quad +
  \quad \lambda J_a(\tilde{{\mathbf x}}, W) \nonumber \\ & = &
  -\frac{1}{N}\sum_{n=1}^N \ln{\sigma_n^y({\mathbf x})} \; + \;
  \lambda \binom{N}{2}^{-1} \sum_{n=1}^N\sum_{m=n+1}^N
  \sigma_n({\tilde{\mathbf x}})^T \sigma_m({\tilde{\mathbf x}})
\label{eq:cost}
\end{eqnarray}
The hyperparameter $\lambda$ controls the trade-off between accuracy
on clean examples and disagreement on perturbed examples. For brevity,
we omit the weight decay regularization. During training, the cost
function is minimized using minibatch stochastic gradient descent. The
cross-entropy term is calculated as a sum over all clean examples in
the minibatch, while the agreement term is calculated over the
perturbed version of the minibatch.

By minimizing both terms simultaneously, the ensemble members are
encouraged to label clean examples accurately while disagreeing on
randomly perturbed examples. This is similar in spirit to a classic
technique in ensemble learning known as {\it negative correlation}
learning~\cite{liu1999ensemble} in which a penalty term is used to
minimize correlation of incorrect predictions on clean data, except
that here we pay no heed to whether the predictions are right or
wrong. Neural networks are often robust to small random noise, which
means that the members may also be penalized for agreeing on the
correct label. This may seem counterintuitive, but the effect is that
the models learn to react differently to perturbed data while still
performing well on clean data, thus encouraging diversity at their
decision boundaries.

\noindent {\bf Detection.} At test time, the outputs of all ensemble
members are combined using a {\it rank voting} mechanism. Each member
assigns the rank $0$ to the label it considers the most likely, rank
$1$ to the second most likely, and so on. For each label, the ranks
are summed across all members, and the smallest label rank is used as
the {\it ensemble disagreement}.

For an ensemble trained using our method, a large ensemble
disagreement is indicative of an input that lies outside of the data
distribution. Correspondingly, we implement a simple rank-based
criterion that rejects a test example as adversarial or outlier if and
only if the ensemble disagreement is above a {\it rank threshold}
hyperparameter $\tau$. We tune $\tau$ to be as low as possible while
having a low false positive rate on clean validation data.

\newcommand{\argmax}{\operatornamewithlimits{argmax}}
\noindent {\bf Classification.} If an input data example is accepted
by the rank thresholding mechanism, one could simply use the label
with the lowest rank sum as the overall prediction of the
ensemble. However, we find in practice that the results are improved
by using a {\it distribution summation} \cite{rokach2009taxonomy} that
selects the classification label as $y^* = \argmax_y \sum_{n=1}^N
\sigma_n^y({\mathbf x})$, where $\sigma_n^y({\mathbf x})$ is the
probability of label $y$ in model $n$. In general, the mechanism for
combining outputs to produce the overall prediction need not be the
same as that for detecting invalid inputs. This could be taken a step
further, such that the entire ensemble is used only for the detection
of invalid inputs, whereas inputs detected as valid are passed to a
separate, perhaps more powerful classifier. We leave this idea for
future investigation.

\vspace{-0.5em}
\section {Experimental Results}
\label{sec:results}

This section answers the following questions:
\begin{enumerate}[labelindent=0pt,leftmargin=*,itemsep=0pt,parsep=0pt, topsep=0pt]
\setlength\itemsep{0pt}
\item How well does the ensemble method of Section~\ref{sec:method}
  defend against known attacks?  We analyze both classification
  accuracy (how well the defense classifies adversarial examples) and
  detection rate (how often the defense detects adversarial
  examples). Detection rate is a measure of the effectiveness of the
  defense when used solely as a detector, assuming that the
  false-positive rate is low. Classification accuracy measures how
  well our ensembles perform when used solely to classify inputs. We
  also measure classification accuracy on accepted inputs (those not
  detected as adversarial), a measure of the effectiveness of
  detection followed by classification.
\item To what degree is the defense susceptible to random noise?  The
  primary metric here is rate of false-positive detection on benign
  noise -- how often the Section~\ref{sec:method} detector classifies
  random noise as adversarial (as this rate increases, the detector
  becomes less useful in deployments).
\end{enumerate}

\noindent{\bf Threat models.}
We measure the effectiveness of attacks in black- and white-box modes,
as well as oblivious-mode in the case of C\&W. White-box attacks have
access to the model, its parameters, and its training data. Black-box
attacks have access to the model and its training data, but not its
parameters. By giving the black-box attacker access to the training
data, we expect a stronger attack than the black-box model
of~\cite{carlini2017adversarial}, in which the attacker and model are
trained on two distinct datasets of comparable size and
quality. Performing additional experiments against weaker black-box
attackers without access to the target model's training data is left
for future work.

\noindent{\bf Attacks.} The fast gradient sign method (FGS) generates
adversarial examples by taking a single linear step from the original
image in the direction of the gradient of the objective function. The
basic iterative method iteratively performs FGS in small steps to
generate more precise examples, subject to a box constraint on
perturbation magnitude. DeepFool attempts to find a minimal
adversarial perturbation by linearizing the discriminant function
around the current perturbed example and iterating until the current
perturbation is sufficient to change the label. The C\&W attack
generates targeted adversarial examples using an objective that
maximizes the margin between the target class logit and the logits of
the other classes, while simultaneously minimizing the perturbation's
$L_2$ norm.

\noindent{\bf Results from targeting $J_e$.} We first evaluate the
defense against attacks which target only the $J_e$ cross-entropy term
of the objective function. In the case of DeepFool and C\&W, as
discriminant function we used the sums of the unnormalized logit
outputs of the ensemble members. As a by-product of maximizing the sum
of logits for a wrong label, the examples inferred by C\&W are
expected to also obtain high ensemble agreement on that label. The
$\kappa$ parameter controlling maximum margin in C\&W is chosen in
each instance to result in comparable mean distortion across attacks.

\begin{table}[tbp]
  \small
  \caption{Accuracy, Detection rate, and mean Distortion for Clean,
    Noise, and Fast Gradient Sign.}
  \label{tableresults1}
  \centering
  \begin{tabular}{*{11}{l}}
    \toprule
    & & \multicolumn{3}{c}{Clean} & \multicolumn{3}{c}{Noise}
    & \multicolumn{3}{c}{FGS} \\
    \cmidrule(l){3-5}
    \cmidrule(l){6-8}
    \cmidrule(l){9-11}
    \multicolumn{2}{c}{Attack} & Acc. & Det. & Dist. & Acc. & Det. & Dist. & Acc. & Det. & Dist. \\
    \midrule
    MNIST & White-box & 98.5/98.8 & 0.9 & 0 & 97.3/99.7 & 30.4 & 1.6 & 16.1/0.0 & 99.8 & 2.8 \\
    & Black-box & & & & & & & 17.5/0.0 & 98.9 & 2.8 \\
    \midrule
    CIFAR-10 & White-box & 81.6/83.9 & 4.6 & 0 & 76.9/83.9 & 39.2 & 0.6 & 10.8/13.6 & 99.8 & 1.7 \\
    & Black-box & & & & & & & 17.5/39.5 & 99.6 & 1.7 \\
    \bottomrule
  \end{tabular}
\end{table}

\begin{table}[tbp]
  \small
  \caption{Accuracy, Detection rate, and mean Distortion for Basic
    iterative, DeepFool, and C\&W.}
  \label{tableresults2}
  \centering
  \begin{tabular}{*{11}{l}}
    \toprule
    & & \multicolumn{3}{c}{Basic iter.} 
    & \multicolumn{3}{c}{DeepFool} & \multicolumn{3}{c}{C\&W} \\
    \cmidrule(l){3-5}
    \cmidrule(l){6-8}
    \cmidrule(l){9-11}
    \multicolumn{2}{c}{Attack} & Acc. & Det. & Dist. & Acc. & Det. & Dist. & Acc. & Det. & Dist. \\
    \midrule
    MNIST & White-box & 13.4/0.0 & 99.7 & 2.4 & 76.0/75.0 & 45.0 & 0.9 & 31.2/95.7 & 97.7 & 2.5 \\
    & Black-box & 22.5/0.1 & 98.9 & 2.4 & 96.7/97.6 & 8.2 & 0.8 & 21.3/33.0 & 76.0 & 2.4 \\
    & Oblivious &       &       &      &       &      &      & 81.6/82.4 & 3.4 & 2.2 \\      
    \midrule
    CIFAR-10 & White-box & 8.0/2.0 & 48.8 & 1.0 & 37.5/38.6 & 42.6 & 0.03 & 7.2/6.3 & 68.1 & 2.3 \\
    & Black-box & 16.9/6.8 & 49.2 & 1.1 & 57.5/68.2 & 36.2 & 0.04 & 9.5/6.9 & 63.6 & 2.4 \\
    & Oblivious &       &       &      &       &      &      &  48.4/50.0 & 6.4 & 1.3 \\
    \bottomrule
  \end{tabular}
\vspace{-2.0em}
\end{table}

We fine-tune the noise hyperparameter $\eta$ of the ensemble method on
validation sets put aside for both MNIST and CIFAR10, yielding
$\eta=0.18$ and $\eta=0.03$, respectively. The selection criterion is
based on false positive rate and detection performance against a
white-box FGS attack. Both ensembles are comprised of $5$ members,
with rank threshold $\tau=2$ for MNIST and $\tau=5$ for CIFAR-10. For
MNIST, each ensemble member is a fully connected network with a single
hidden layer (128 neurons). For CIFAR-10, we use a standard
convolutional network drawn from TensorFlow~\cite{tf-cifar10}.

Tables \ref{tableresults1} and \ref{tableresults2} show the
performance of the ensembles on unperturbed (Clean), noisy (Noise),
and adversarial inputs (FGS, Basic iterative, DeepFool, C\&W), in
terms of classification accuracy (Acc) on all/accepted adversarial
examples, detection rate (Det), and the mean $\mathrm{L}_2$ distortion
(Dist) of the adversarial examples. For the Noise column, noisy
examples are generated by applying benign random noise to the clean
data set. For FGS and Basic iterative, $\epsilon = 0.1$ and $\epsilon
= 0.03$ were used for MNIST and CIFAR-10, respectively. Random noise
examples were generated with $\eta = 0.1$ and $\eta = 0.02$, using the
same algorithm that was used during training. Higher values of $\eta$
cause significant misclassification in the CIFAR-10 case, so such
noise is not benign.

We see reasonable classification accuracy and false-positive rate on
clean data for both ensembles. Sensitivity to random noise is
increased, but classification accuracy on accepted noisy examples is
significantly higher than usual. FGS and basic iterative are
completely blocked with the exception of the basic iterative attack on
CIFAR-10, which is moderately successful. The most successful attack
is DeepFool, causing significant white-box classification error with
only 42.6\% detection and very small mean distortion. However, we find
that the same attack reduces the classification accuracy of an
undefended MNIST classifier to 1.32\%, and CIFAR-10 to 12.02\%. The
oblivious C\&W attack might also be considered successful, as it
achieves similar classification error with very low detection rate,
but the mean distortion is much higher. There appears to be a
trade-off between classification accuracy and detection rate. Further
analysis is required to determine the cause of this trend.

The C\&W attack was recently shown~\cite{carlini2017adversarial} to be
quite effective at generating adversarial examples, obtaining success
rates close to 100\% against ten detection methods. By setting
$\kappa=250$ in the C\&W attack, we were able to perform an oblivious
attack against our ensemble method which caused 9.6\% classification
accuracy and only 7.0\% detection rate. However, the mean distortion
was 3.3, which we consider to be very high. 
intuition about the detection method, so further analysis is required
to fully understand its consequences.

\noindent{\bf Results from targeting $J_e + \lambda J_a$.} The FGS and
Basic iterative attacks can be used to target $J_e$ and $J_a$
simultaneously, in order to both fool the ensemble and bypass the
defense. Tables \ref{tablemnistcombined} and \ref{tablecifarcombined}
show results of white- and black-box attacks that target both terms of
the ensemble objective function simultaneously. Maximizing the first
term causes misclassification, while maximizing the second causes high
agreement. The expectation in this case is that most members of the
ensemble agree on an incorrect label, thereby successfully causing
misclassification while avoiding detection. The scaling parameter
$\lambda$ can be tuned to trade off between classification accuracy
and detection.

We choose values for $\lambda$ that illustrate its effect on the
outcome of the attacks. The terms $J_e$ and $J_a$ are weighted equally
when $\lambda = 1$. As $\lambda$ decreases, less weight is given to
$J_a$. Again, we observe a trade-off between classification accuracy
on adversarial examples and detection. FGS is largely ineffective, but
the basic iterative method achieves greater success when $\lambda$ is
chosen properly, resulting in 26.4\% accuracy and only 27.1\%
detection in the white-box setting when $\lambda=0.27$.

\begin{table}[tbp]
  \small
  \caption{Accuracy, Detection rate, and mean Distortion for MNIST
    attack on $J_e + \lambda J_a$.}
  \label{tablemnistcombined}
  \centering
  \begin{tabular}{*{12}{l}}
    \toprule
    & & \multicolumn{3}{c}{$\lambda=4$} 
    & \multicolumn{3}{c}{$\lambda=1$} & \multicolumn{3}{c}{$\lambda=0.25$} \\
    \cmidrule(l){3-5}
    \cmidrule(l){6-8}
    \cmidrule(l){9-11}
    \multicolumn{2}{c}{Attack} & Acc. & Det. & Dist. & Acc. & Det. & Dist. & Acc. & Det. & Dist. \\
    \midrule
    FGS & White-box & 89.9/98.5 & 94.2 & 2.8 & 88.2/98.4 & 94.5 & 2.8 & 19.1/0.0 & 99.8 & 2.8 \\
        & Black-box & 97.0/98.8 & 70.8 & 2.8 & 95.2/98.9 & 71.2 & 2.8 & 18.3/1.7 & 98.8 & 2.8 \\
    Basic iter. & White-box & 96.9/97.5 & 56.6 & 1.9 & 45.0/91.3 & 89.6 & 2.0 & 15.2/3.9 & 99.5 & 2.4 \\
    & Black-box & 96.9/98.5 & \justify{6.9} & 1.4 & 93.2/98.7 & 12.5 & 1.4 & 24.2/12.4 & 98.6 & 2.4 \\
    \bottomrule
  \end{tabular}
\end{table}

\begin{table}[tbp]
  \small
  \caption{Accuracy, Detection rate, and mean Distortion for CIFAR-10
    attack on $J_e + \lambda J_a$.}
  \label{tablecifarcombined}
  \centering
  \begin{tabular}{*{12}{l}}
    \toprule
    & & \multicolumn{3}{c}{$\lambda=1$} 
    & \multicolumn{3}{c}{$\lambda=0.27$} & \multicolumn{3}{c}{$\lambda=0.25$} \\
    \cmidrule(l){3-5}
    \cmidrule(l){6-8}
    \cmidrule(l){9-11}
    \multicolumn{2}{c}{Attack} & Acc. & Det. & Dist. & Acc. & Det. & Dist. & Acc. & Det. & Dist. \\
    \midrule
    FGS & White-box & 35.9/80.6 & 99.4 & 1.7 & 20.6/43.5 & 99.5 & 1.7 & 11.7/10.8 & 99.6 & 1.7 \\
    & Black-box & 33.2/74.5 & 99.1 & 1.7 & 20.1/46.4 & 99.4 & 1.7 & 18.8/37.5 & 99.5 & 1.7 \\
    Basic iter. & White-box & 57.0/59.4 & \justify{7.9} & 1.0 & 26.4/32.2 & 27.1 & 1.0 & \justify{8.9}/3.3 & 47.2 & 1.0 \\
    & Black-box & 53.2/56.3 & \justify{9.3} & 1.1 & 22.2/16.7 & 43.2 & 1.1 & 17.4/7.7 & 48.2 & 1.1 \\
    \bottomrule
  \end{tabular}
\vspace{-0.5em}
\end{table}

\section{Conclusion and Future Work}
\label{conclusion}
We propose a new ensemble learning method\footnote{The source code is
  available at https://github.com/bagnalla/ensemble\_detect\_adv.} for
detecting adversarial examples that is both attack-agnostic and
computationally inexpensive. We evaluate its effectiveness against
four known attacks (FGS, basic iterative, DeepFool and C\&W) in
oblivious, black-box, and white-box settings. In future work, we plan
to incorporate adversarial re-training into our ensemble method, and
to experiment with separate models for detection and classification.

\bibliographystyle{plain}
\bibliography{paper}

\end{document}